\documentclass[runningheads]{llncs}
\usepackage{graphicx}
\usepackage{amssymb}
\usepackage{makecell}
\usepackage{multirow}

\usepackage{soul}
\usepackage{url}
\usepackage[hidelinks]{hyperref}
\usepackage[utf8]{inputenc}
\usepackage[small]{caption}
\usepackage{graphicx}
\usepackage{amsmath}
\usepackage{multirow}
\usepackage{booktabs}
\usepackage{adjustbox}
\usepackage{placeins}
\usepackage[misc]{ifsym}
\usepackage[ruled,vlined]{algorithm2e}

\SetKwInput{KwInput}{Input}                % Set the Input
\SetKwInput{KwOutput}{Output}              % set the Output

\begin{document}
\title{Tackling Inter-Class Similarity and Intra-Class Variance for Microscopic Image-based Classification}
\titlerunning{Inter-Class Similarity and Intra-Class Variance in Classification}
\author{Aishwarya Venkataramanan\inst{1,2,3}\orcidID{0000-0002-6100-0034}[\Letter] \and 
Martin Laviale\inst{1,3}\orcidID{0000-0002-9719-7158} \and
C\'ecile Figus\inst{1} \and
Philippe Usseglio-Polatera\inst{1}\orcidID{0000-0001-7980-7509} \and
C\'edric Pradalier\inst{2,3}\orcidID{0000-0002-1746-2733}}

\institute{Laboratoire Interdisciplinaire des Environnements Continentaux, Universit\'e de Lorraine, France \email{venkatar1@univ-lorraine.fr}
 \and GeorgiaTech Lorraine- International Research Lab Georgia Tech - CNRS IRL 2958, France
 \and LTER- "Zone Atelier Moselle", France
}
\authorrunning{A. Venkataramanan et al.}
%\institute{Princeton University, Princeton NJ 08544, USA \and
%Springer Heidelberg, Tiergartenstr. 17, 69121 Heidelberg, Germany
\maketitle              % typeset the header of the contribution
\begin{abstract}
Automatic classification of aquatic microorganisms is based on the morphological features extracted from individual images. The current works on their classification do not consider the inter-class similarity and intra-class variance that causes misclassification. We are particularly interested in the case where variance within a class occurs due to discrete visual changes in microscopic images. In this paper, we propose to account for it by partitioning the classes with high variance based on the visual features. Our algorithm automatically decides the optimal number of sub-classes to be created and consider each of them as a separate class for training. This way, the network learns finer-grained visual features. Our experiments on two databases of freshwater benthic diatoms and marine plankton show that our method can outperform the state-of-the-art approaches for classification of these aquatic microorganisms.

\keywords{Micro-organisms classification  \and Automatic clustering \and Intra-class variance \and  Inter-class similarity }
\end{abstract}
\section{Introduction}
Micro-organisms are key components of aquatic ecosystems and analysing their distribution is a central question in aquatic ecology. This often requires manual identification of organisms by a human expert using a microscope. Organisms are grouped into taxa based on morphological features which can present a huge diversity. The classification task often turns out to be time-consuming, tedious and sometimes requires a high-level of expertise. Thus, methods are being developed to automate the process~\cite{du1999diatom,zheng2017automatic}. Over the years, the automatic classification has evolved from traditional hand-crafted methods~\cite{bueno2017automated,zhao2010binary} to deep-learning-based ones~\cite{lumini2020deep,pastore2020annotation}. While the methods using deep learning have shown improved performance, we show that they are not sufficient to learn fine-grained visual features for reliable classification~\cite{pedraza2017automated}.

\begin{figure}[t]
    \centering
    \begin{tabular}{cccccc}
        %\multicolumn{3}{c}{\bf Inter-class similarity}&\multicolumn{3}{c}{\bf Intra-class variance}\\ 
        \includegraphics[height=3.0cm,width=0.10\linewidth]{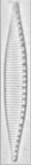} & 
        \includegraphics[height=3.0cm,width=0.10\linewidth]{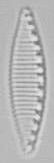}&
        \includegraphics[height=3.0cm,width=0.10\linewidth]{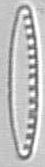}&
        \hspace{2cm}
        \includegraphics[height=3.0cm,width=0.10\linewidth]{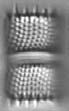} & 
        \includegraphics[height=3.0cm,width=0.10\linewidth]{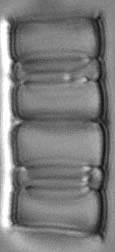} &
        \includegraphics[width=0.25\linewidth]{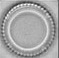} \\
        \multicolumn{3}{c}{\bf (a)}&\multicolumn{3}{c}{\bf (b)}\\       
        
    \end{tabular}
    \caption{Examples of diatom images usually misclassified due to high inter-class similarity (a) or intra-class variance (b). (a) Three different diatom species of the genera \textit{Nitzschia}: \textit{N. soratensis}, \textit{N. subacicularis} and \textit{N. costei}, from left to right. (b) A single diatom species \textit{Aulacoseira pusilla} seen from the side or from the top, from left to right.
    }
    
    \label{fig:diatoms}
\end{figure}

For this study, we consider the classification task of two typical aquatic microorganisms: freshwater benthic diatoms and marine plankton. Diatoms are unicellular micro-algae characterized by a highly ornamented silicified exoskeleton. Plankton includes a group of organisms drifting or floating in the water column, encompassing a large range of microbes, algae and larvae. Automated classification of these microscopic images is challenging notably due to two aspects: inter-class similarity and intra-class variance. Inter-class similarity occurs when objects belonging to different classes have visually similar appearance due to minute variations in the morphological features. Intra-class variance is when some objects belonging to the same class have drastically different appearances. This is prevalent in microscopic images due to the restrictions in view-points from which the images are acquired. Diatoms are typically imaged on permanent microscope slides: samples are treated chemically in order to remove organic materials and the diatom suspension is allowed to settle out, which restricts each organism to lie either on its side or top. While in planktons, after appropriate sub-sampling by size-selective filtration, images of live samples are acquired using a submersible imaging flow-cytometer, which ensures that every processed organisms are analysed individually. Indeed due to imaging constraints, the images acquired using these microscopy methods result in discrete view-points. Examples of diatoms with inter-class similarity and intra-class variance are shown in Fig.~\ref{fig:diatoms}. Fig.~\ref{fig:diatoms}(a) shows the images of diatoms from three different classes that have similar appearance and are often confused by the classification network. Fig.~\ref{fig:diatoms}(b) shows the images from a single class, but taken from the side and top view. Here, the network fails to identify all of them as belonging to the same class.  
%Our diatom dataset: freshwater diatoms, but mainly benthic (few planktonic that can sediment

An intuitive approach to tackle this problem would be to separate the feature embeddings of different classes by a distance so that the network can better differentiate them. This is known as metric-learning~\cite{chen2017beyond,Schroff_2015_CVPR} and the principle behind it is to bring closer the feature embeddings of the objects belonging to the same class while pushing apart the embeddings of the objects belonging to different classes. 
This is effective for inter-class similarity, but does not impose any constraints on the intra-class variance. Hence, in this paper, we propose an algorithm that automatically clusters the instances within the high variance classes. Then the generated clusters are considered as independent classes while training. The clustering is based on the learned visual features and so, each cluster contains similar feature embeddings. In this way, the variance within each class is reduced, and the network can learn fine-grained features specific to them. Our technical contribution is that our algorithm automatically chooses the classes to be clustered and the optimal number of clusters to be generated. Finally, to handle the problem of inter-class similarity, we use triplet loss to increase the distance between the inter-class features. Experimental results on a diatom and plankton dataset show that by reducing the impact of intra-class variance and inter-class similarity, the network learns finer-grained features that improves the classification performance. 

Our contributions in this paper can be summarised as follows:
(1) We propose a method to address the problem of inter-class similarity and intra-class variance in fine-grained visual classification, particularly in the setting where the images within a class have contrasting appearance changes.
(2)We apply our method to a real-world problem of diatom and plankton classification and show that our method has improved performance over the existing state-of-the-art approaches.

\section{Related Work}
Early methods of plankton and diatom classification relied on hand-crafted methods to extract features for classification. \cite{zheng2017automatic} uses multiple kernel learning to perform plankton classification. \cite{du1999diatom} is a pilot study on using image processing techniques to perform automatic detection of diatoms. \cite{bueno2017automated} analyses different morphological and statistical descriptors to classify the diatoms. 
Although the hand-crafted based methods have been proven to be successful in identification, it is time-consuming to choose the appropriate features. Thus, the recent focus is on using deep learning to extract features automatically.

\cite{py2016plankton} develops a deep network model to exploit the translational and rotational symmetry in plankton images. \cite{lumini2020deep} compares the performance of various deep neural networks for plankton and coral classification. \cite{pastore2020annotation} develops a combination of unsupervised and supervised learning to classify plankton images. However, none of these works consider the influence of intra-class variance and inter-class similarity on the classification performance.

Few methods have been proposed in the computer vision literature to handle intra-class variance and inter-class similarity for visual recognition. \cite{gadermayr2015dealing} uses split-and-merge to handle high intra-image and intra-class variations for celiac disease diagnosis. ~\cite{cacheux2019modeling} considers the instances contributing to high intra-class variance as outliers. They use triplet loss with a weighting scheme where each instance is given a weight based on how representative they are of their class.~\cite{pilarczyk2019intra} uses a Hadamard layer to minimise the intra-class variance. The one closely related to ours is ~\cite{em2017incorporating}, where they cluster the instances within each class into a set of pre-defined sub-classes (K classes) using K-Means clustering. They calculate two sets of triplet losses: one for the broader class instances and the other for the sub-class instances. The drawback with this approach for our application is that since K-Means is applied to all the classes and K is pre-defined, it could result in over or under-clustering. 
Contrary to their approach, our algorithm uses X-Means~\cite{pelleg2000x} and decides the classes to be clustered and the optimal number of clusters during training. X-Means is a variant of K-Means that can automatically decide the number of clusters to be created based on certain conditions such as Akaike information criterion (AIC) or Bayesian information criterion (BIC) scores. 

\section{Method} \label{Method}
%In addition to the clustering, we use triplet loss to increase the distance between the feature embeddings of each class and so can handle the classes with high inter-class similarity.
The proposed method considers the two challenges of aquatic microorganism classification: (1) inter-class similarity and (2) intra-class variance.

\subsection{Inter-class Similarity}
Inter-class similarity results in the network learning similar feature embeddings for the classes with similar-looking images. During inference, this causes confusion in differentiating between them. Along with the cross-entropy loss commonly used for classification networks, we use triplet loss to separate the feature embeddings between each class. The triplet loss tries to minimise the distance between the objects belonging to the same class while maximising the distance between the objects belonging to different classes in the feature space. This can be formulated as follows: Let $x_a$, $x_p$ and $x_n$ be the anchor, positive and negative image. Here $x_a$ and $x_p$ belong to the same class while $x_n$ is from a different class. Let $f$ be the function to obtain the feature embeddings of the images. The triplet loss is given by 
\begin{equation}
    L_{triplet} = max\{||f(x_a)-f(x_p)||-||f(x_a)-f(x_n)||+\alpha , 0\}
\end{equation}
where $\alpha$ is the margin to separate the positive and negative images in the feature space. Our final loss is the sum of the cross-entropy loss and the triplet loss.
\begin{equation}
    L_{total} = L_{cross-entropy} + L_{triplet}
\end{equation}

\subsection{Intra-class variance}
%When capturing images using a microscope, the orientation of the organism being observed can be different and this can result in drastic changes in the visual appearance. Apart from this, there could be deformations in the organisms or they can occur in tightly knitted groups that can contribute to visual changes. This results in high intra-class variance within that class. 
The classification network fails to identify all the images as belonging to the same class when there is a high variance of data within the class. Our proposed method clusters the classes with high variance into sub-groups and consider these sub-groups as independent classes for classification.

We use X-Means since it automatically decides the clusters to be generated. However, it can sometimes generate non-optimal number of clusters. When there is over-clustering, there will be overlap of visual features between two or more clusters. Considering these clusters as independent classes further aggravates the problem of inter-class similarity. Having a smaller number of clusters still doesn't solve the intra-class variance problem. Thus, the goal of our method is to find the optimal number of clusters to minimize both the inter-class similarity and intra-class variance. X-Means uses a parameter for the upper limit to the number of allowed clusters, and always generates clusters that are less than or equal to this value. Our algorithm decides this parameter for each class so that the optimal number of clusters is generated.

\begin{figure}[t]
    \centering
    \begin{tabular}{cccc|ccccc}
    \rotatebox{90}{\scriptsize Cluster1} &
    \includegraphics[width=0.12\linewidth]{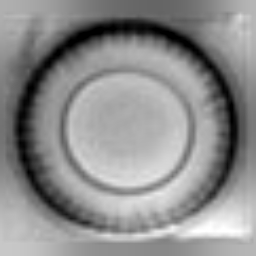} &
    \includegraphics[width=0.12\linewidth]{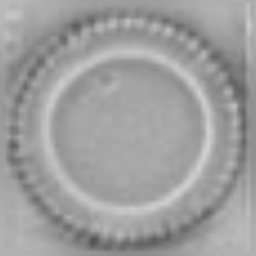} & 
    \includegraphics[width=0.12\linewidth]{AUPU_clusters/1/IDF_AUPU_058027.png} &
    &\rotatebox{90}{\scriptsize Cluster 2} &
    \includegraphics[width=0.12\linewidth]{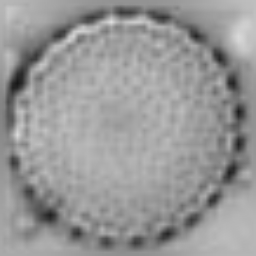} &
    \includegraphics[width=0.12\linewidth]{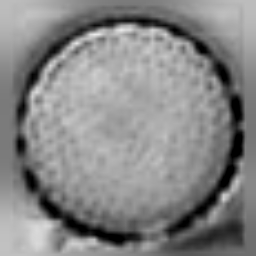} &
    \includegraphics[width=0.12\linewidth]{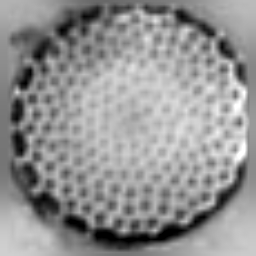} \\
    \rotatebox{90}{\scriptsize Cluster 3} &
    \includegraphics[width=0.12\linewidth]{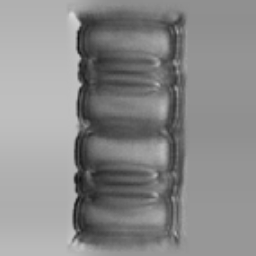} &
    \includegraphics[width=0.12\linewidth]{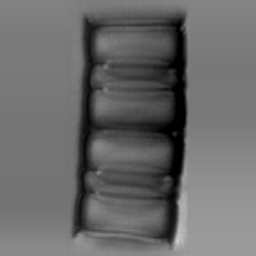} &
    \includegraphics[width=0.12\linewidth]{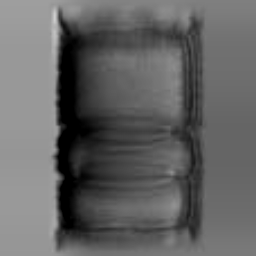} &
    &\rotatebox{90}{\scriptsize Cluster 4} &
    \includegraphics[width=0.12\linewidth]{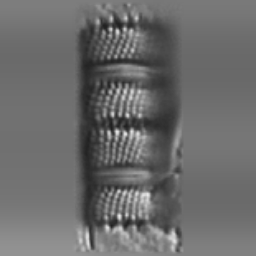} &
    \includegraphics[width=0.12\linewidth]{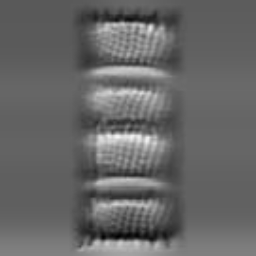} &
    \includegraphics[width=0.12\linewidth]{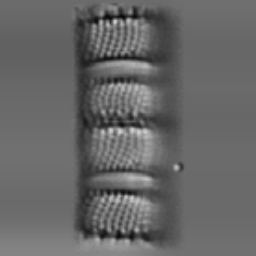} \\
    \end{tabular}
    \caption{Clusters generated based on the learned visual features for the diatom class \textit{Aulacoseira pusilla}.  }
    \label{fig:aupu_clusters}
\end{figure}

\subsection{Training algorithm}
For every class in the data, we define a parameter \texttt{num-allowed-clusters}, which is the upper limit to the number of clusters that can be generated for that class. As the training progresses, \texttt{num-allowed-clusters} is adapted until it reaches the optimal value. There is a threshold to the maximum value that can be attained by \texttt{num-allowed-clusters} which is \texttt{max-clusters}.
The overall training can be divided into the following steps and the process is repeated until the training converges:
    1) Extract feature embeddings for the training images; 
    2) Cluster using X-Means;
    3) Train the classification encoder network; 
    4) Perform validation and obtain the normalised confusion matrix; 
    5) Adjust \texttt{num-allowed-clusters} based on the confusion matrix.

%Hence during the course of training, we adapt \texttt{num-allowed-clusters} for each class, which is the upper limit to the     %One should note that \texttt{max-allowed-clusters} is the upper limit to the number of clusters generated and X-Means can also generate lesser number of clusters.

In order to generate the clusters, we extract the feature embeddings of the images from the last fully-connected layer of a classification network pre-trained on the ImageNet dataset. X-Means uses these feature embeddings to cluster the images. At the beginning of training, we initialise \texttt{num-allowed-clusters} to 1 for all the classes, which is the same as training a standard classifier. However as the training progresses, we adapt \texttt{num-allowed-clusters} and X-Means generates clusters based on this value. These clusters are then considered as independent classes while training the network. Fig.~\ref{fig:aupu_clusters} shows an example of clusters generated for the diatom class \textit{Aulacoseira pusilla}. The diatoms have been clustered based on the difference in view-point and the porosity. Since they are considered as independent classes, the network learns these intricate details which helps in fine-grained classification. 

\begin{algorithm}[h]
\DontPrintSemicolon
  
  \KwInput{Training and validation data, \texttt{max-clusters}, \texttt{confusion-threshold}. Initialise number of clusters for each class with 1 and flag for each class with 0. }
  \KwOutput{Trained model}
  %\KwData{Testing set $x$}
  
  \While{Not Converged}
      {
   		Train the encoder network;\\
   		Perform validation and get the normalised confusion matrix;\\
   		Extract features of the images in the training data;\\
   		\For{each class}
   		{\If{false negative $>$ \texttt{confusion-threshold} and flag=0}
   		{ Increase \texttt{num-allowed-clusters}. If the number of clusters reach the \texttt{max-allowed-clusters} set the flag to 1.
   		}
   		\ElseIf{false negative $>$ \texttt{confusion-threshold} and flag=1}
   		{Decrease \texttt{num-allowed-clusters}. If the number of clusters reach 1 then set the flag to 0. }
   		Generate the clusters and pseudo-labels using X-Means.
   		}

   } \label{alg:clustering}
\caption{Training with clustering}
\end{algorithm}
After every epoch of training, we obtain the normalised confusion matrix. When the false-negative for a particular class is above \texttt{confusion-threshold}, we increment \texttt{num-allowed-clusters} for that class. For our application, the classification certainty of each class is important and so we use false-negatives to optimize \texttt{num-allowed-clusters}.
As the training progresses and when the false-negatives is greater than the \texttt{confusion-threshold}, we increment \texttt{num-allowed} \texttt{-clusters} until it reaches \texttt{max-clusters}. %Once the maximum value is reached and the false-negative still remains above the threshold, this could be because \texttt{num-allowed-clusters} has overshot the optimal value of the number of clusters and so it causes over-clustering. 
During the initial stages of the training, when the feature embeddings learned by the network are not fully refined, the false negatives are relatively high for many of the classes. Thus the false negatives of these classes will be above the \texttt{confusion-threshold} and the number of clusters may overshoot the optimal value. Thus we start decrementing \texttt{num-allowed-clusters} until it reaches 1. This adaptation continues until the optimal value is reached. 
The above process is repeated for every epoch until the training converges. The summary of our algorithm is given in Algorithm 1.

\section{Experiments}

\subsection{Datasets}
%Our proposed method deals with intra-class variance when the instances within the classes appear as discrete sub-spaces. This is a common scenario in microscopic images due to the restricted view-points from which the images are captured.
We apply our method on two datasets:

\textbf{Diatom Dataset} The diatom dataset consists of individual images of diatom from three different public taxonomic atlases~\cite{DREAL3}. The dataset contains a total of 166 classes with a total of 9895 images.

\textbf{WHOI-Plankton Dataset~\cite{orenstein2015whoi}}
The WHOI plankton dataset consists of 3.4 million images spread across 70 classes. We considered only those classes that have at least 50 images. Finally we obtained 38 classes and a total of 26612 images.

All the images were padded and resized to size $256\times256$. We used K-Fold cross-validation with K=5.

\subsection{Baselines}
To evaluate our method we perform experiments on the following baselines: 
\begin{enumerate}
    \item \textbf{Standard Classification} - We use a state-of-the-art classification network pre-trained on ImageNet dataset and fine-tune to our dataset. 
    \item \textbf{Classification with triplet loss} - Along with the cross-entropy loss, we use the triplet loss. This method is done to study the impact of inter-class similarity on the classification performance.
    \item \textbf{Classification with clustering} - This is our proposed clustering method, but using only the cross-entropy loss for classification. This is used to study the impact of intra-class variance on the classification performance.
    \item \textbf{Classification with clustering and triplet loss} - This is our proposed method to minimise the impact of both the inter-class similarity and the intra-class variance.
    \item \textbf{GS-TRS~\cite{em2017incorporating}} - This method uses K-Means to divide each class into K clusters and uses triplet loss for inter-cluster and inter-class objects.
\end{enumerate}
We perform our experiments on two classification model architectures: ResNet50 and EfficientNet.
 
\subsection{Evaluation metrics}
We evaluate the different approaches using the standard metrics used in classification, namely the classification accuracy, precision, recall, F-Score. A higher value of these metrics indicates a better performance. 
Additionally, we also calculate the variance of the per-class false-negatives and false-positives. The variance gives us a measure of how consistent the classification is and so a lower value is preferred.
%when there are some classes with perfect classification and some with a high percentage of imperfect classification, the variance will be high in these cases.   

\subsection{Implementation Details}
We use Adam optimizer and the learning rate is 0.0002 and we use a batch size of 128. The output feature embedding dimension from the network is 256. We trained our networks on GeForce GTX 1080 with 12 GB RAM. The value of \texttt{max-allowed-clusters} is set to 5. Our \texttt{confusion-threshold} is set to 0.3 based on hyper-parameter search.

\section{Results}

\begin{table}[!htb]
  \caption{Quantitative metrics for classification on the diatom dataset.} 
  \centering
  %\begin{minipage}{0.1\linewidth}
  %\begin{adjustbox}{center, width=0.85\linewidth} 
  \begin{tabular}{|c|c|c|c|c|c|c|c|}\hline
   
   \multirow{2}{*}{Architecture}&\multirow{2}{*}{Method}&\multirow{2}{*}{Accuracy}& \multirow{2}{*}{Recall} &\multirow{2}{*}{Precision}&\multirow{2}{*}{F Score}& \multicolumn{2}{c|}{Variance} \\
        &&&&&& FN & FP\\\hline\hline
        \multirow{5}{*}{ResNet50}&{Std. Classification}&{94.24}&{93.54}&{93.98}&{92.85}&{0.036}&{0.049}\\\cline{2-8}
        &{Classification+triplet loss}&{96.06}&{95.86}&{95.97}&{95.33}&{0.022}&{0.023}\\\cline{2-8}
        &{Classification+clustering}&{96.57}&{96.53}&{96.61}&{96.15}&{0.020}&{0.014}\\\cline{2-8}
        &{Ours}&{\textbf{97.37}}&{\textbf{97.95}}&{\textbf{97.97}}&{\textbf{97.79}}&{\textbf{0.0059}}&{\textbf{0.0071}}\\\cline{2-8}
        &{GS-TRS~\cite{em2017incorporating}}&{93.23}&{93.03}&{93.44}&{92.66}&{0.018}&{0.018}\\\hline\hline
        
        \multirow{5}{*}{EfficientNet}&{Std. Classification}&{95.31}&{94.92}&{95.10}&{94.82}&{0.031}&{0.041}\\\cline{2-8}
        &{Classification+triplet loss}&{96.67}&{96.52}&{96.68}&{96.50}&{0.021}&{0.021}\\\cline{2-8}
        &{Classification+clustering}&{96.76}&{96.61}&{96.11}&{96.67}&{0.019}&{0.016}\\\cline{2-8}
        &{Ours}&{\textbf{97.22}}&{\textbf{96.64}}&{\textbf{97.30}}&{\textbf{96.69}}&{\textbf{0.0047}}&{\textbf{0.0053}}\\\cline{2-8}
        &{GS-TRS~\cite{em2017incorporating}}&{93.60}&{92.96}&{93.27}&{93.43}&{0.020}&{0.024}\\\hline

  \end{tabular}
  %\end{adjustbox}
  %\end{minipage}
  \label{tab:diatom}
\end{table}

\subsection{Diatom dataset}
Table~\ref{tab:diatom} shows the quantitative metrics for the diatom classification. Our results show that using both clustering and triplet loss consistently outperforms the other methods. One interesting conclusion from the results is that classification with clustering performs better than classification with triplet loss. This means that intra-class variance has a higher impact on the classification performance than the inter-class similarity. Finally, optimizing both the inter-class similarity and intra-class variance further improves the performance of the network. GS-TRS~\cite{em2017incorporating} is not well-suited for this application because the number of clusters generated are not optimal and over or under-clustering deteriorates the performance.

\begin{figure}[!h]
    \centering
    \begin{tabular}{cc}
    \includegraphics[width=0.5\linewidth]{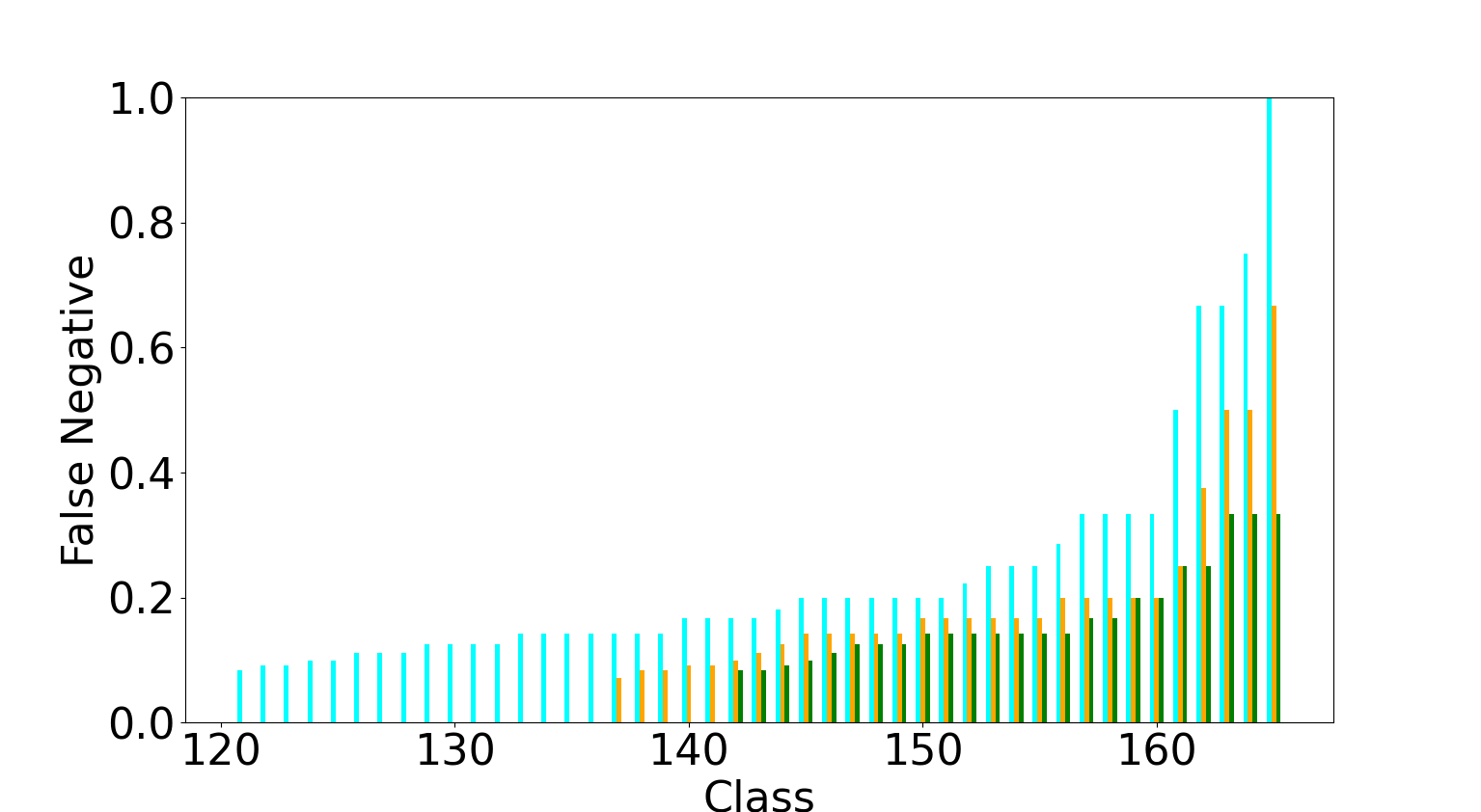} &
    \includegraphics[width=0.5\linewidth]{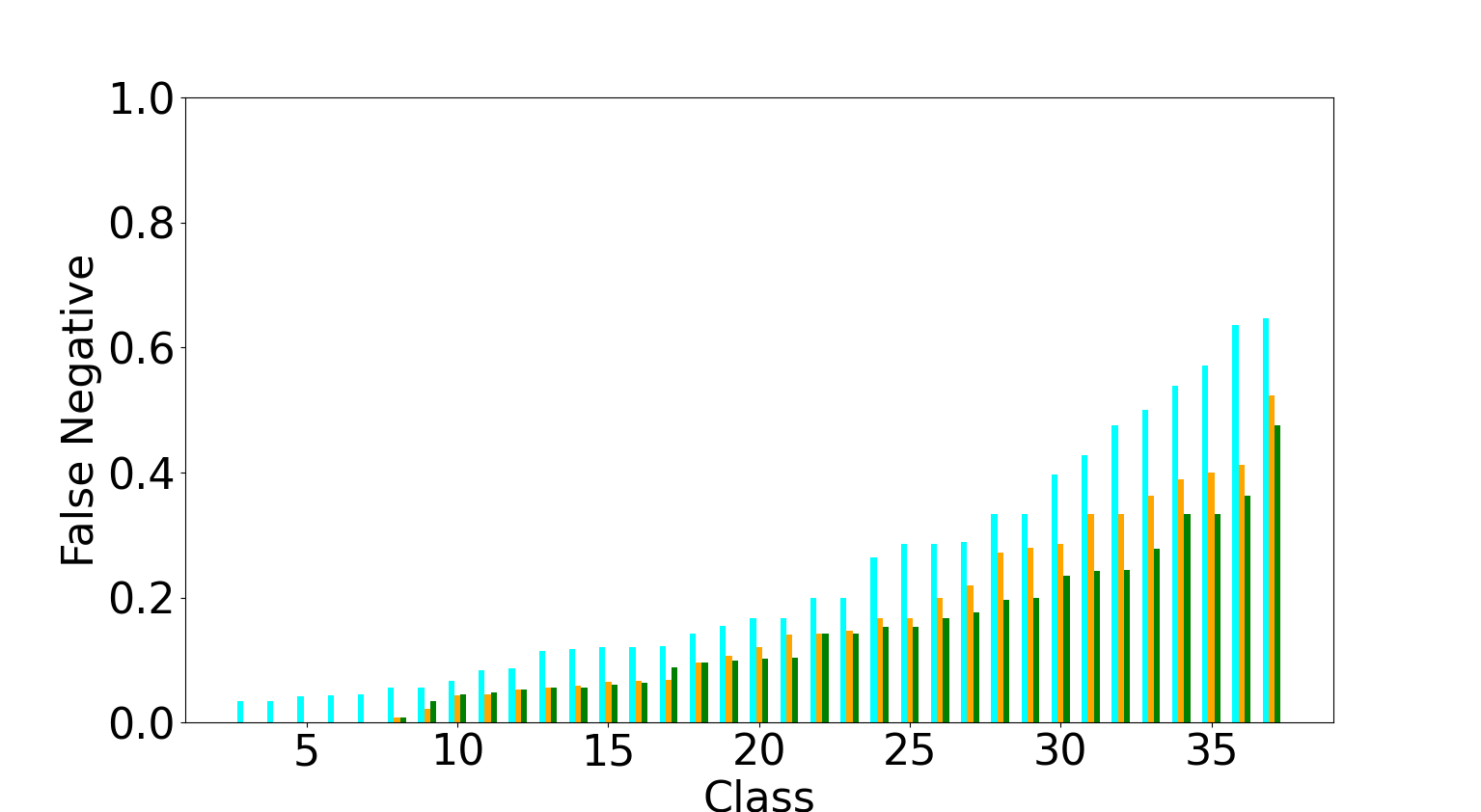} \\
    {(a)} & {(b)}
    \end{tabular}
    \caption{Per-class false negatives of standard classification(cyan), classification with clustering(orange) and classification with clustering and triplet loss(green) for (a) diatom dataset and (b) WHOI-Plankton dataset with EfficientNet. \textit{Note: The classes with 0 false negatives are not shown here.}} 
    \label{fig:fn_overlay}
\end{figure}

For automatic classification of aquatic microorganisms, the certainty of prediction of a class is important which means that the false negatives should be minimal. Fig.~\ref{fig:fn_overlay}(a) shows a zoomed version of the false negatives with 3 methods: standard classification, classification with clustering, and classification with clustering and triplet loss overlaid onto a single graph. From the plots, 120 classes are perfectly classified when using a state-of-the-art classifier whereas 140 classes are perfectly classified when using clustering with triplet loss. Also, the overall magnitude of the false negatives is reduced when using clustering and triplet loss than when compared to the other methods. This is a significant improvement since the network can reliably be used to identify a larger number of classes than before.

\subsection{WHOI-Plankton Dataset}

\begin{table}[t]
  \caption{Quantitative metrics for classification on the WHOI-plankton dataset.} 
  \centering
  %\begin{minipage}{0.1\linewidth}
  %\begin{adjustbox}{center, width=0.85\linewidth} 
  \begin{tabular}{|c|c|c|c|c|c|c|c|}\hline
   
   \multirow{2}{*}{Architecture}&\multirow{2}{*}{Method}&\multirow{2}{*}{Accuracy}& \multirow{2}{*}{Recall} &\multirow{2}{*}{Precision}&\multirow{2}{*}{F Score}& \multicolumn{2}{c|}{Variance} \\
        &&&&&& FN & FP\\\hline\hline
        \multirow{5}{*}{ResNet50}&{Std. Classification}&{88.54}&{84.90}&{85.82}&{84.71}&{0.022}&{0.136}\\\cline{2-8}
        &{Classification+triplet loss}&{89.25}&{83.90}&{85.39}&{84.63}&{0.034}&{0.084}\\\cline{2-8}
        &{Classification+clustering}&{88.17}&{\textbf{87.49}}&{83.25}&{84.52}&{\textbf{0.013}}&{0.095}\\\cline{2-8}
        &{Ours}&{\textbf{89.48}}&{85.64}&{86.54}&{\textbf{85.50}}&{0.017}&{\textbf{0.034}}\\\cline{2-8}
        &{GS-TRS~\cite{em2017incorporating}}&{87.53}&{78.75}&{\textbf{86.67}}&{80.81}&{0.035}&{0.156}\\\hline\hline
        
        \multirow{5}{*}{EfficientNet}&{Std. Classification}&{88.82}&{86.51}&{82.21}&{81.67}&{0.030}&{0.125}\\\cline{2-8}
        &{Classification+triplet loss}&{88.99}&{84.31}&{84.23}&{83.78}&{0.041}&{0.079}\\\cline{2-8}
        &{Classification+clustering}&{88.71}&{85.96}&{81.65}&{81.49}&{0.015}&{0.103}\\\cline{2-8}
        &{Ours}&{\textbf{90.53}}&{\textbf{88.91}}&{\textbf{87.13}}&{\textbf{83.86}}&{\textbf{0.013}}&{\textbf{0.024}}\\\cline{2-8}
        &{GS-TRS~\cite{em2017incorporating}}&{87.66}&{82.25}&{82.35}&{82.03}&{0.033}&{0.147}\\\hline

  \end{tabular}
  %\end{adjustbox}
  %\end{minipage}
  \label{tab:whoi}
\end{table}
Table~\ref{tab:whoi} shows the quantitative metrics for classification on the WHOI-Plankton dataset. Similar to the diatoms, the clustering along with triplet loss outperforms the other methods. %However, here the classification with triplet loss has a better performance than the classification with clustering. This is because the inter-class similarity has a higher impact on classification performance than the intra-class variance for this dataset.  
Fig.~\ref{fig:fn_overlay}(b) shows the overlay plot of the false negatives. In contrast to three classes that were perfectly identified by the state-of-the-art classifier, clustering and triplet loss improves it to seven perfectly identified classes. One could observe from Fig.~\ref{fig:fn_overlay}(b) that when using only clustering and when using clustering along with triplet loss, the false negative magnitude does not change much. This is due to the relatively lower number of classes in the WHOI-plankton dataset, which reduces the impact of the inter-class similarity.  

\section{Conclusion}
In this paper, we proposed a method to tackle the inter-class similarity and intra-class variance due to discrete image subsets, which is commonly found in microscopic images. Our method automatically identifies the classes to be clustered and the optimal number of clusters to be generated. Then these clusters are considered as independent classes while training a classification network. Finally, to deal with the inter-class similarity, we use triplet loss to separate out the features between each class. Using this approach, the network was able to learn finer-grained features that improved the classification performance. This was validated using quantitative metrics on a diatom and a plankton dataset. %Our future work would focus on generalising this method to continuous-subspaces that are commonly found in macroscopic images.  

\bibliographystyle{splncs04}
\bibliography{biblio}

\end{document}